\documentclass[letterpaper, 10 pt, conference]{ieeeconf}
\IEEEoverridecommandlockouts
\usepackage{times}
\usepackage{epsfig}
\usepackage{graphicx}
\usepackage{amsmath}
\usepackage{amssymb}
\usepackage{booktabs}
\usepackage[nolist,nohyperlinks]{acronym}
\usepackage{caption}
\usepackage{subcaption}
\usepackage{parskip}
\usepackage{units}
\usepackage[hidelinks]{hyperref}

\begin{document}

%%%%%%%%% TITLE
\title{\LARGE \bf \vspace{0.7em}
CalQNet -- Detection of Calibration Quality for Life-Long Stereo Camera Setups
}

\author{Jiapeng Zhong$^\ast$, Zheyu Ye$^\ast$, Andrei Cramariuc, Florian Tschopp, Jen Jen Chung\\
Roland Siegwart, Cesar Cadena%
\thanks{$^\ast$Authors contributed equally to this work.}%
\thanks{All the authors of this paper are affiliated with the Autonomous Systems Lab, ETH Z\"{u}rich, {\tt\small \{jzhong, zheyye, crandrei, ftschopp, chungj, rsiegwart, cesarc\}@ethz.ch}.}%
}

\maketitle
\thispagestyle{empty}
\pagestyle{empty}

%%%%%%%%% ABSTRACT
\begin{abstract}
    Many mobile robotic platforms rely on an accurate knowledge of the extrinsic calibration parameters, especially systems performing visual stereo matching.
    Although a number of accurate stereo camera calibration methods have been developed, which provide good initial ``factory'' calibrations, the determined parameters can lose their validity over time as the sensors are exposed to environmental conditions and external effects.
    Thus, on autonomous platforms on-board diagnostic methods for an early detection of the need to repeat calibration procedures have the potential to prevent critical failures of crucial systems, such as state estimation or obstacle detection.
    In this work, we present a novel data-driven method to estimate the calibration quality and detect discrepancies between the original calibration and the current system state for stereo camera systems. 
    The framework consists of a novel dataset generation pipeline to train CalQNet, a deep convolutional neural network.
    CalQNet can estimate the calibration quality using a new metric that approximates the degree of miscalibration in stereo setups.
    We show the framework's ability to predict from a single stereo frame if a state-of-the-art stereo-visual odometry system will diverge due to a degraded calibration in two real-world experiments.
\end{abstract}

%%%%%%%%% BODY TEXT
\section{Introduction}
In the emerging industries of autonomous robotic systems, such as micro aerial vehicles and self-driving cars, sensor calibration plays an essential role and is especially important for stereo visual sensor setups~\cite{strecha2008benchmarking}.
In order to estimate an accurate calibration, a sophisticated procedure~\cite{Roth1987AnCalibration} is usually performed beforehand, and the obtained parameters are kept for the life-cycle of the product.
The most accurate and commonly used calibration methods involve the use of a known target board and require the user to perform expert calibration procedures~\cite{Zhang2000ACalibration,abraham2005fish,rehder2016extending}.
However, the originally determined calibration parameters can lose their validity over time, due to environmental transients and external disturbances, such as mechanical shocks.
Such changes will lead to unavoidable biases and errors in the processes that utilize the sensor data, for example state estimation or obstacle detection algorithms which can lead to critical failures of vital systems of the vehicle.
Yet, as more and more autonomous platforms and applications emerge, the capability of users to perform sophisticated calibration task to update the calibration parameters should not be an expected requirement.
Online target-less calibration methods also exist, but are computationally very expensive and less precise~\cite{furukawa2009accurate,dang2009continuous,zhang1996motion,horaud2000stereo}.
Thus, fault detection~\cite{Visinsky1994RoboticSurvey} for warning the user about the need for recalibration in a timely manner, can greatly alleviate the dangers posed by degraded calibrations~\cite{Ni2009SensorTypes}, regardless of which method for recalibration would be used.

In this work, we propose a data-driven method for detection of miscalibration for stereo camera setups, to bypass the need to recalibrate in order to check the validity of the current parameters.
We propose a method in which for each specific stereo setup, a network is tailor trained to predict when a recalibration is necessary for the current setup.

Using a learning-based approach poses its own set of challenges, among which firstly is the lack of a publicly available large-scale dataset, that would feature life-long setups of stereo visual sensors where the calibration parameters degrade over time.
To tackle this issue we adopt a similar semi-synthesized data generation approach as proposed in~\cite{Cramariuc2020LearningDetection}.
The second challenge is that the degree of miscalibration in the extrinsics of a stereo camera is difficult to quantify.
A classic approach would be to calculate a degree of miscalibration by measuring the average distance of features to corresponding epipolar lines~\cite{faugeras1993three}.
However, this procedure relies on feature matching for each stereo pair, which performs poorly at high disturbance.
We address these challenges and provide the following key contributions:
\begin{itemize}
    \item A novel metric, \ac{WODE}, quantifying extrinsic miscalibration of stereo camera setups.
    \item A data generation pipeline to train a \ac{CNN} called CalQNet, that predicts from a single stereo frame whether the platform is miscalibrated.
    \item Experiments on calibration quality prediction performance and use cases in two real world experiments using ORB-SLAM2~\cite{Mur-Artal2017ORB-SLAM2:Cameras}.
\end{itemize}

\section{Related Work}
\label{sec:rel-work}
One of the most common ways to evaluate the monocular and stereo camera calibration is using reprojection errors of known keypoints in the scene, e.g. by using calibration targets or fiducial markers~\cite{Hartley2004MultipleVision}.
In such controlled environments and under expert handling,  accurate sensor calibrations can be obtained using a variety of methods~\cite{Zhang2000ACalibration,abraham2005fish,rehder2016extending,Atcheson2010}.

In addition to methods based on known target observations, target-less calibration methods have also been investigated, also due to their potential to be used online during the life-cycle of the sensor.
Such calibration methods~\cite{furukawa2009accurate,dang2009continuous,zhang1996motion,horaud2000stereo} can be used during online operation and in unknown environments, since they rely on a joint optimization of tracked feature locations and camera parameters.
Similarly, more robustness can be added by including external sensors as well into the problem, as proposed in~\cite{Zhang2018AOdometry,Schneider2019Observability-AwareEstimation}, where an inertial measurement unit is used in addition to a camera.
However, all the aforementioned online calibration methods are computationally expensive, produce less accurate calibrations and potentially require additional external sensors.
Due to these disadvantages, these methods are impractical for the purpose of confirming the validity of the current calibration parameters in the regular use of stereo camera systems.

Other online methods that do not directly deal with the calibration, but perform sensor fault detection, rely on correlating information from redundant sensors~\cite{Mendoza2012,Roumeliotis1998SensorRobot,Sundvall2006}.
In addition to the computational requirements these methods are more costly since they require multiple sensors to integrate and work independently.
Fault detection can also be indirectly done by performing sanity-checks on outputs of algorithms that rely on good calibrations~\cite{bar-shalom}.
In these cases it becomes ambiguous whether the reason for the fault was bad calibration, the type of motion or the current environment. 
Furthermore, for many systems, performing tasks with potentially failing algorithms can pose a severe safety threat.
Many optimization based odometry methods such as ORB-SLAM2~\cite{Mur-Artal2017ORB-SLAM2:Cameras} have been shown to be able to function with inaccurate calibrations, but at a significant degradation in performance.

In~\cite{Cramariuc2020LearningDetection} the authors propose a learning-based algorithm for detecting when the intrinsic parameters of monocular cameras have diverged from the original calibration.
This proves the capability and applicability of \acp{CNN} to detect mismatches between the observed environment and the undistorted image.
However, the same method can not be directly applied to the sensor extrinsics problem, as the process of image undistortion is fundamentally different from stereo image rectification.

\section{Methodology}
\label{sec:approach}

\begin{figure*}[t]
    \vspace{0.7em}
    \centering
    \subfloat{\includegraphics[width=0.75\linewidth]{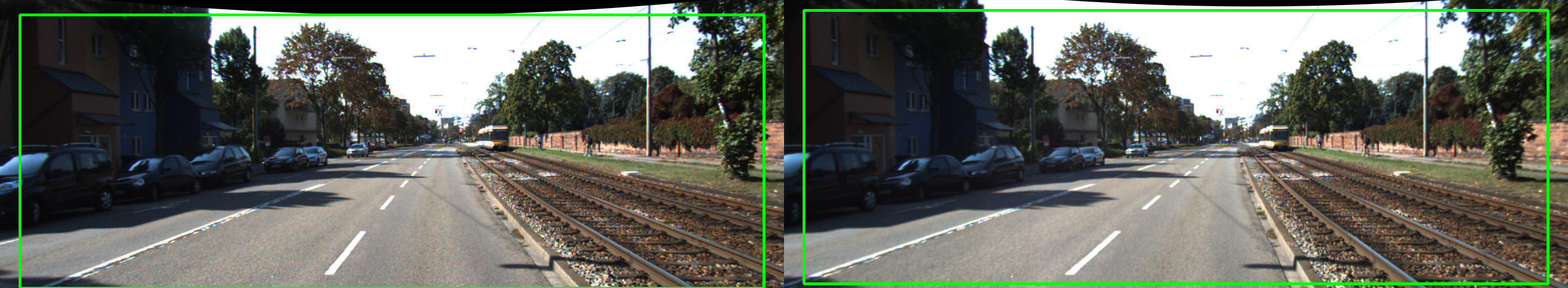}}
    \hfill
    \subfloat{\includegraphics[width=0.75\linewidth]{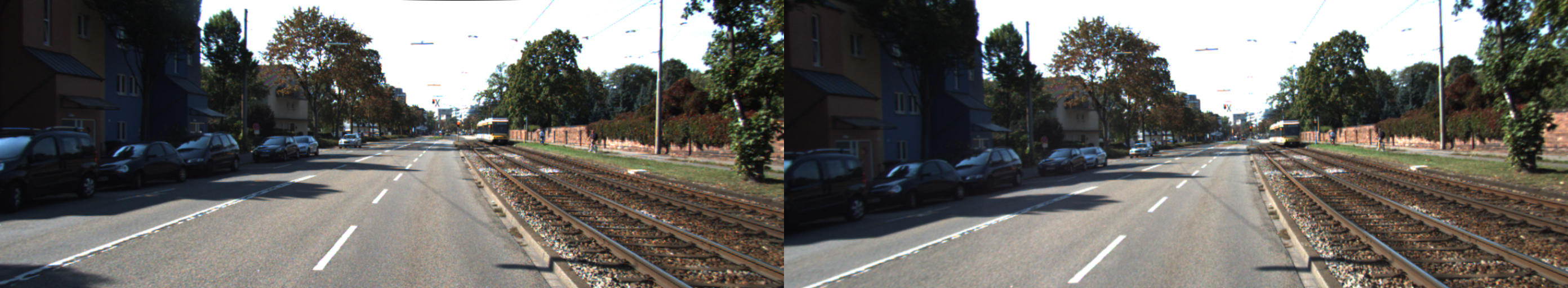}}
    \caption{
    Example of a rectified stereo pair with true calibration parameters from the KITTI dataset~\cite{Geiger2013}.
    \textbf{Top:} Validity mask $\mathcal{R}$ (green) corresponding to the largest rectangle containing only valid pixels with same aspect ratio as the original image.
    \textbf{Bottom:} Valid region after cropping and resizing to original image size as final input into the network}
    \label{fig:rectified_true}
\end{figure*}

\begin{figure*}
    \centering
    \begin{subfigure}[b]{0.39\linewidth}
        \centering
        \includegraphics[width=\linewidth]{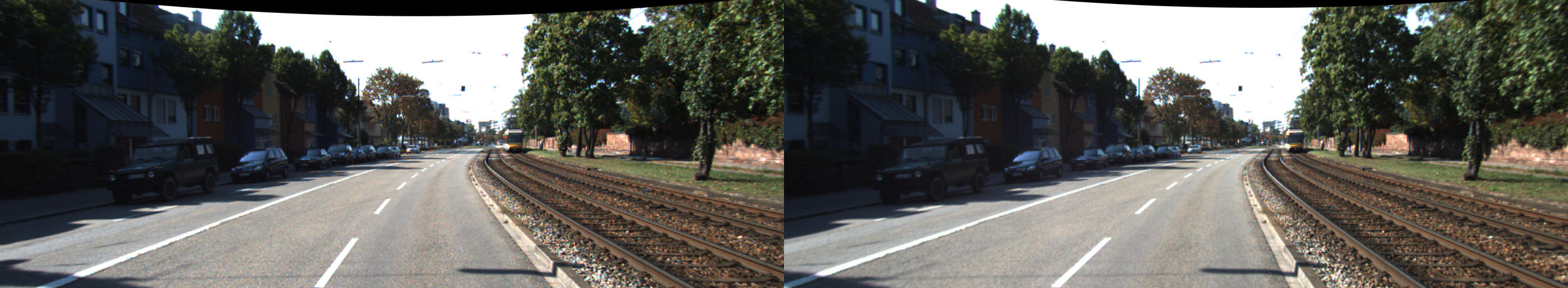}
        \label{fig:dtx}
    \end{subfigure}
    \begin{subfigure}[b]{0.39\linewidth}
        \centering
        \includegraphics[width=\linewidth]{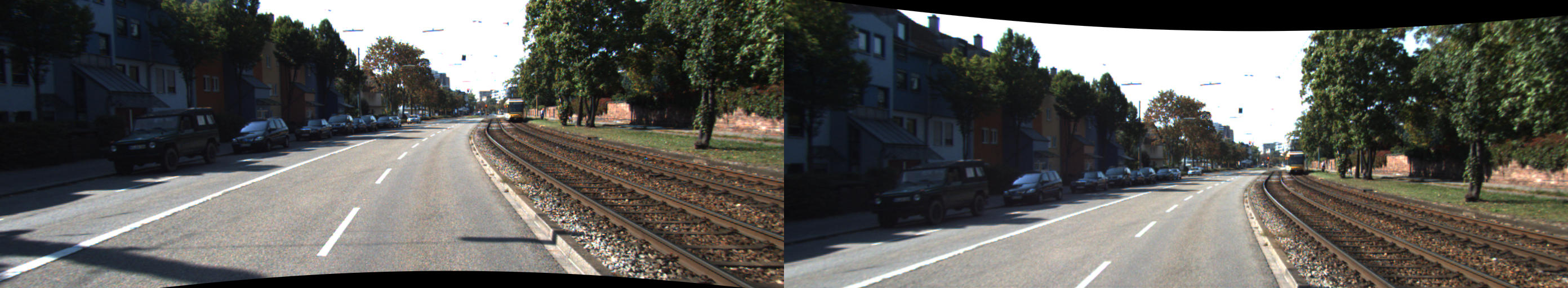}
        \label{fig:drx}
    \end{subfigure}

    \begin{subfigure}[b]{0.39\linewidth}
        \centering
        \includegraphics[width=\linewidth]{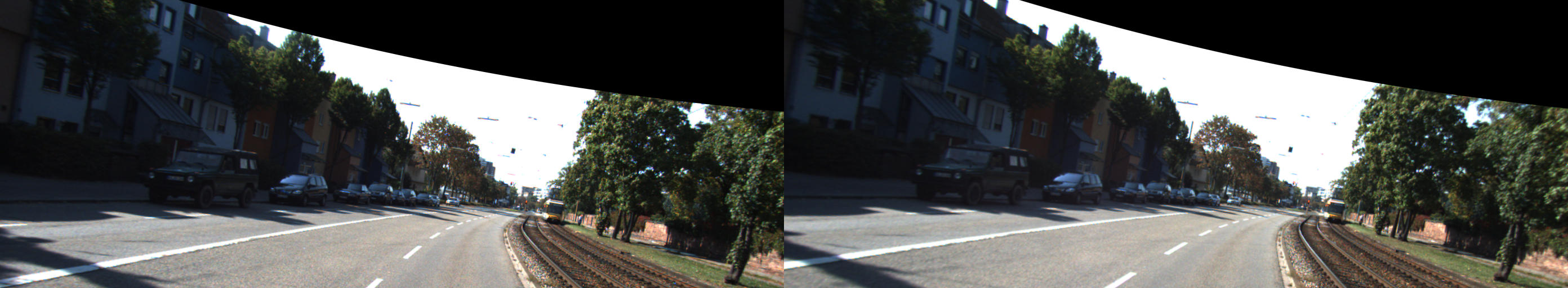}
        \label{fig:five over x}
     \end{subfigure}
    \begin{subfigure}[b]{0.39\linewidth}
        \centering
        \includegraphics[width=\linewidth]{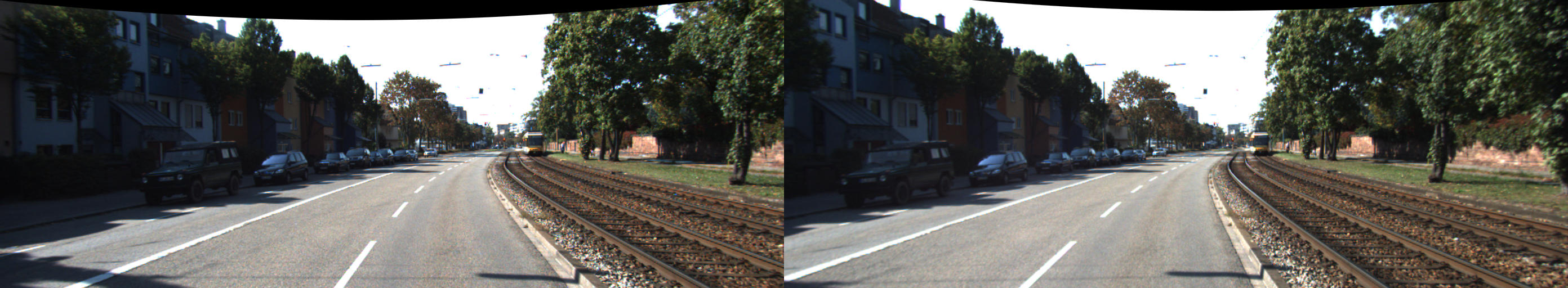}
        \label{fig:dty}
    \end{subfigure}

    \begin{subfigure}[b]{0.39\linewidth}
        \centering
        \includegraphics[width=\linewidth]{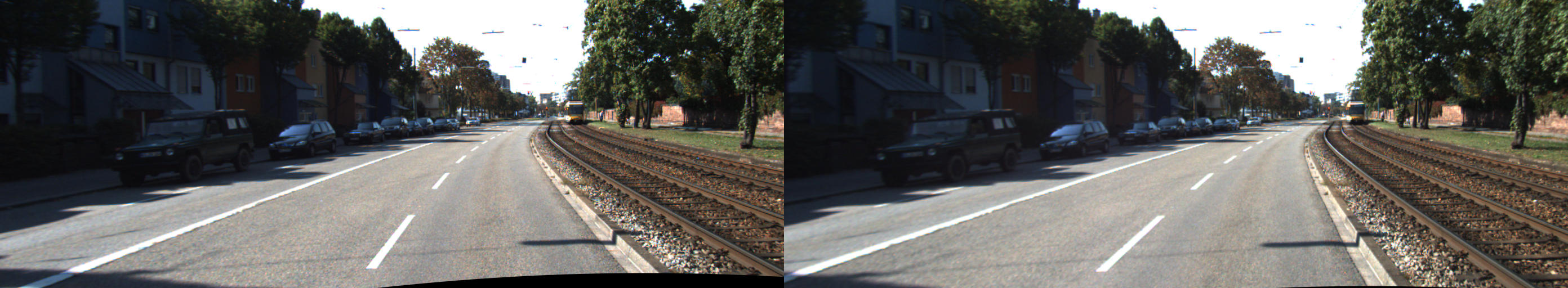}
        \label{fig:dtz}
    \end{subfigure}
    \begin{subfigure}[b]{0.39\linewidth}
        \centering
        \includegraphics[width=\linewidth]{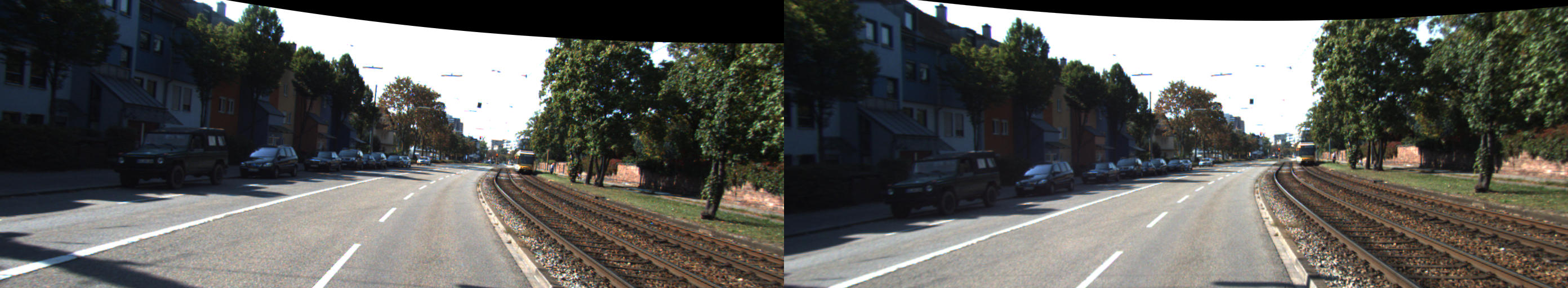}
        \label{fig:drz}
    \end{subfigure}

    \caption{Examples of stereo rectifications with the same amount of disturbance. For different \acp{dof}, the disturbance affects image rectification differently. \textbf{Left}: disturbance of $\unit[0.1]{m}$ in $x$, $y$, and $z$, respectively. \textbf{Right}: disturbance of $\unit[0.1]{rad}$ in $\alpha$, $\beta$, and $\gamma$, respectively.
    }
    \label{fig:disturbance_effect}
\end{figure*}

In this section, our contributions are presented in detail.
The dataset generation pipeline is explained in Section~\ref{ssec:dataset}. In Section~\ref{ssec:metric}, our novel metric for miscalibration is defined.
Finally, the neural network CalQNet and its training are detailed in Section~\ref{ssec:network}.

\subsection{Dataset Generation}
\label{ssec:dataset}
In a real-life scenario pre-calibrated parameters are usually stored and remain untouched, and what changes during calibration drift is the physical sensor setup.
Thus, the most straightforward procedure to create a dataset to represent such a scenario, is to manually vary the extrinsics between sensors from the nominal one and take measurements with the disturbed setup.
However, in this process re-calibration would be required each time the setup is perturbed in order to understand the deviation.
This is time-consuming and impractical, especially since the procedure also needs to be repeated for each individual sensor setup.
This repetition is necessary because we do not propose a generic solution for detecting when any sensor has a bad calibration, but instead we show that it is possible to fine-tune CalQNet to detect when a specific sensor and its current calibration are no longer valid.
Thus, we adopt the strategy of semi-synthesization, introduced in~\cite{Cramariuc2020LearningDetection}.
This includes modifying the calibration parameters by adding randomly generated disturbance and re-rectifying the stereo pairs accordingly.
This is equivalent to the visual effect on a setup, where the ``true calibration'' is the disturbed one and the amount of miscalibration is the disturbance added.

In our stereo dataset generation pipeline, we assume each camera has been correctly calibrated, i.e., the intrinsics and distortion coefficients are known and remain valid throughout the process of stereo rectification, and only the extrinsics are considered variable.
The true extrinsic transformation (translation, rotation) is denoted as $\Theta = \{x,y,z,\alpha,\beta,\gamma\}$.
For each pair of raw camera images $I_{left}$ and $I_{right}$, a rectification map $M' = f(\Theta)$\footnote{This mapping also depends on the camera intrinsics and distortions. However, in our notation this effect is excluded by our assumption that they remain fixed.} can be calculated to obtain the rectified images $I'_{left}$ and $I'_{right}$~\cite{Hartley2004MultipleVision}.
Due to the projective transformation and tangential undistortion process, the resulting rectified images have different shapes from the original images.
Nevertheless, the use of a \ac{CNN} requires a unified input size.
Therefore, in order to account for all these restrictions, we define a validity mask $\mathcal{R}$, which corresponds to the largest rectangular region in the rectified images containing only valid pixels, while retaining the aspect ratio of the original images.
The cropped areas are further resized to the size of the raw images, filling the empty pixels with bilinear interpolation, to get the final sample images $\hat{I}_{left}$ and $\hat{I}_{right}$.
An example of the stereo rectification process and validity mask is shown in Figure~\ref{fig:rectified_true}.

We use the calibration parameters $\Theta^\ast$ provided by the dataset as ground truth, and the resulting image pairs and rectification maps as $\hat{I}^\ast$ and $\hat{M}^\ast$, respectively.
To obtain samples of miscalibration, a random vector $d\in \mathbb{R}^6$ is sampled uniformly between $[-1.5d_{thr}, 1.5d_{thr}]$, where $d_{thr}$ is a manually set threshold so that the application is still expected to function.
We train by sampling a higher interval so that CalQNet can learn to predict miscalibration cases where the target application is no longer expected to work.
In our case, $d_{thr}$ is chosen according to the performance of ORB-SLAM2, which will be discussed in detail in Section~\ref{sec:exp}.
This process allows us to generate an arbitrarily large dataset for training a \ac{CNN} with a relatively small number of raw images and one valid calibration.

\subsection{Metrics for Quantifying Miscalibration}
\label{ssec:metric}

As described in Section~\ref{ssec:dataset}, stereo rectification with accurate intrinsics is a transformation parameterized by the extrinsic calibration parameters.
Due to the continuous nature of these parameters, one can apply arbitrarily small disturbance to the calibration.
Furthermore, the same amount of disturbance applied to different directions has a different effect on the mapping, as shown in Figure~\ref{fig:disturbance_effect}.

Therefore, we propose adding weight terms to quantify the different effects on the transformation.
This is achieved by capturing the influence of each degree of freedom $i$ in $\Theta$.
During the stereo rectification, image planes of stereo cameras are rotated with corresponding projective transformation matrices $H_{left}(d)$ and $H_{right}(d)$ to make both images co-planar, which are dependent of the extrinsic parameters.
Thus, we calculated a weight $w_i$ based on how much the original plane is rotated differently from the true calibration, i.e.
\begin{multline}
w_{i}(d_i) = \sum \left\Vert Rodrigues(H^{-1}_{left,i}(d_i)\cdot{}\hat{H}^\ast_{left}) \right\Vert_2\\
        + \sum \left\Vert Rodrigues(H^{-1}_{right,i}(d_i)\cdot{}\hat{H}^\ast_{right}) \right\Vert_2,
        \label{eq:weights}
\end{multline}
where $i$ is the index of image pairs, and $\hat{H}^\ast_{left}$ and $\hat{H}^\ast_{right}$ are the projection matrices calculated using the true transformation vector.
The inverse of the matrices calculated with disturbed parameters are multiplied with the ones from true parameters and then converted to rotation vectors.
As the calculation of such projective transformations is dependent on both the translation and rotation between the two cameras, this process quantitatively captures the relative difference between the effects on the resulting rectified image plane for the same amount of disturbance in different \ac{dof}.
Finally, the overall effect \ac{WODE} $\delta$ is calculated as the weighted sum of disturbances in all directions
\begin{equation}
    \delta = \sum_{i\in{DoF(6)}}\left\Vert{} w_{i}(d_i) \cdot{}d_{i}\right\Vert .
\end{equation}

\begin{figure*}[t]
    \centering
    \includegraphics[width=0.7\linewidth]{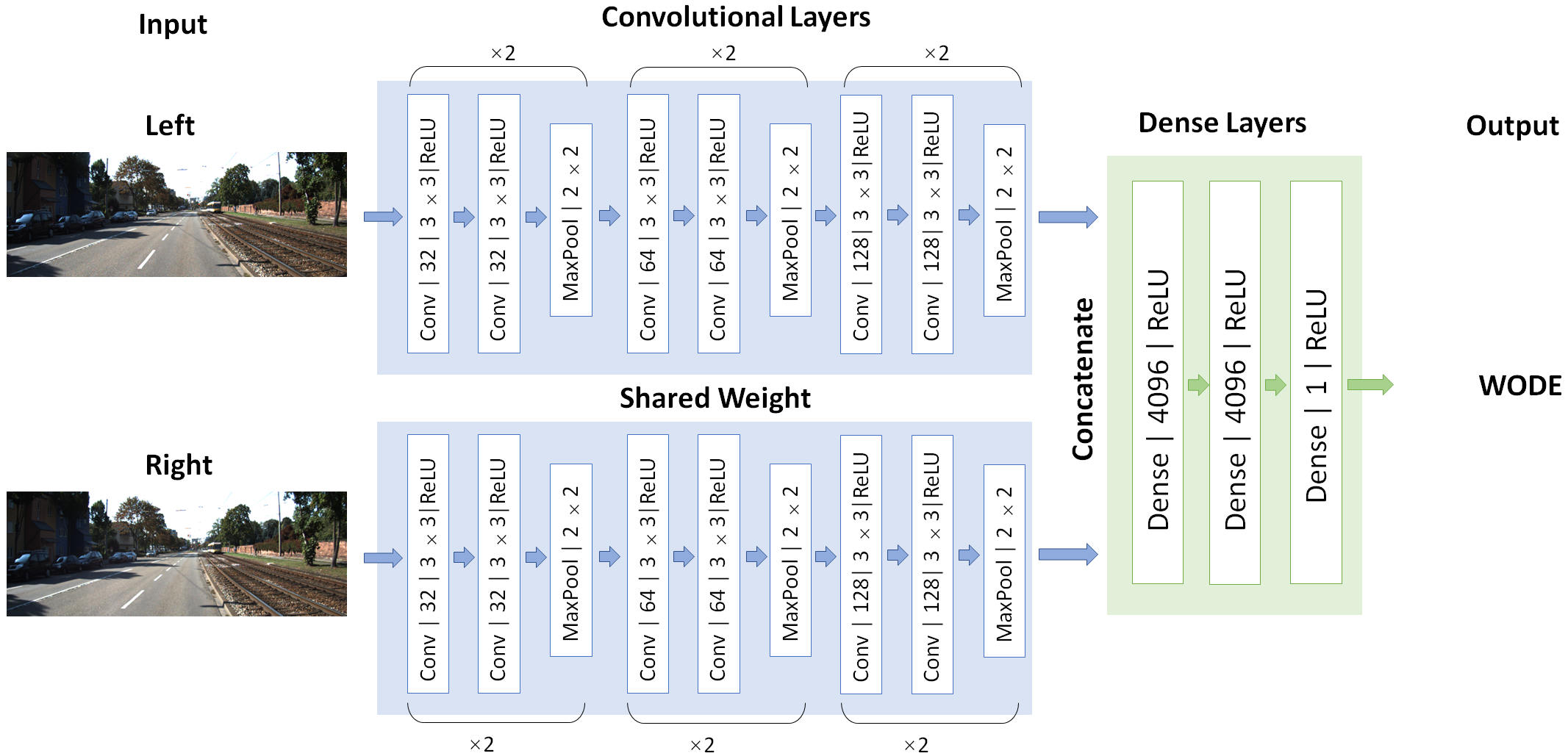}
    \caption{Network architecture of CalQNet to predict \ac{WODE}. The input is a pair of rectified stereo images which are first processed in a Siamese network and then concatenated through fully connected layers to predict the calibration quality. }
    \label{fig:network_architecture_stereo}
\end{figure*}

For training, $\delta$ is further normalized in order to achieve ${\bar{\delta}=1}$ for the fully disturbed case, i.e. disturbing every \ac{dof} with $d_{thr}$
\begin{equation}
    \bar{\delta} = \frac{\delta}{\delta_{thr}} = \frac{\sum_{i\in{DoF(6)}}\left\Vert{} w_{i}(d_i) \cdot{}d_{i}\right\Vert}
            {\sum_{i\in{DoF(6)}}\left\Vert{}w_{i,thr} \cdot{} d_{thr}\right\Vert},
\end{equation}
where $\delta_{thr}$ is the normalization factor and $w_{i,thr}$ is the weight calculated with $d = d_{thr}$ in Equation~\eqref{eq:weights} on the corresponding \ac{dof}.
We choose to normalize the distribution of \ac{WODE} for training to avoid numerical instability in the network, when $d_{thr}$ is chosen to be very small.
Furthermore, it makes the output more interpretable with respect to the expected functioning of the target application.

\subsection{Network for Detecting Stereo Miscalibration}
\label{ssec:network}
The architecture of the \ac{WODE} prediction network CalQNet for stereo cameras is
presented in Figure~\ref{fig:network_architecture_stereo}.
The base network is adopted from~\cite{Cramariuc2020LearningDetection}, and transformed into a \textit{Siamese} architecture~\cite{BROMLEY1993SIGNATURENETWORK}, where two branches of \acp{CNN} share the same structure and weights.
Images pairs are rectified and fed separately through the convolutional backbone and the outputs are concatenated together as an input to a dense network for final predictions of the calibration validity.
During training, mean squared error loss is applied to the error between the networks \ac{WODE} prediction and the ground truth \ac{WODE}.
The loss is optimized using the ADAM optimizer~\cite{Kingma2015Adam:Optimization} with a reductive learning rate (halved every 20 epochs).
Dropout~\cite{Srivastava2014Dropout:Overfitting} is applied to avoid over-fitting and the network parameters are initialized with Xavier's initialization method~\cite{Glorot2010UnderstandingNetworks}. 
\section{Experiments and Evaluations}
\label{sec:exp}
In this section, we evaluate our trained model and metric on two publicly available datasets. The results of evaluating the trained models are discussed in Section~\ref{ssec: performance-kitti}. The evaluation of \ac{WODE} as a metric is detailed in Section~\ref{ssec:eval-wode}. Finally, in Section~\ref{sec:eval-orbslam} we show how \ac{WODE} and CalQNet can be used in conjunction with an odometry system.

\subsection{Miscalibration Detection with a Network}
\label{ssec: performance-kitti}

To evaluate the performance of CalQNet in predicting \ac{WODE}, we show results on the widely used KITTI~\cite{Geiger2013} and EuRoC~\cite{Burri2016TheDatasets} datasets, which feature various stereo camera setups.
For KITTI we only use the sequences from September 26, 2011 and perform a random split into $\unit[80]{\%}$ for training data and the rest for testing.
A similar split is performed for all the sequences in EuRoC.
We add uniformly sampled translational and rotational disturbance to the ground truth calibration as detailed in Section~\ref{ssec:dataset}.
The thresholds for sampling are $d_{thr, \text{KITTI}} = 0.05$ and $d_{thr, \text{EuRoC}} = 0.025$, expressed in meters or radians.
The threshold values were determined empirically based on the range of values within which an odometry system might still work.
The two datasets have different threshold due to the very different stereo setups with which they were recorded.
The choice of $d$ results in maximal \ac{WODE} values of around $\bar{\delta}_{max}\sim 2$.

\begin{figure}[t]
    \centering
    \includegraphics[width=0.49\linewidth]{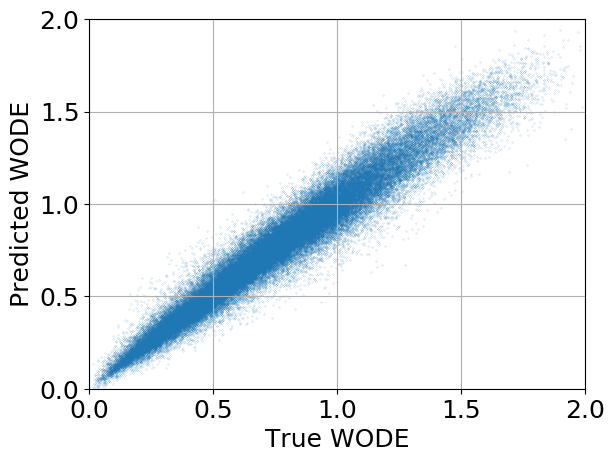}
    \includegraphics[width=0.49\linewidth]{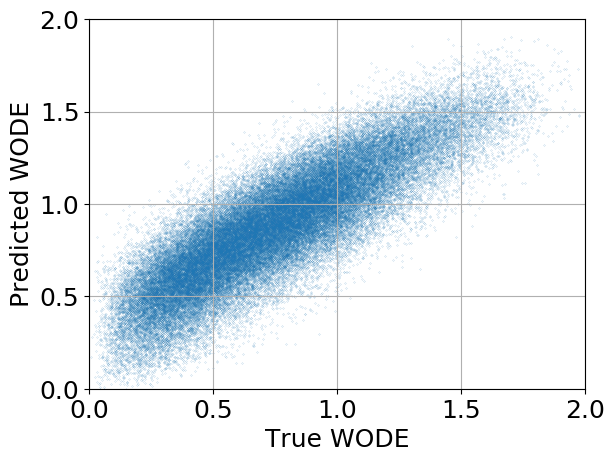}
    \caption{Network performance on KITTI dataset ($\delta_{thr} = 0.0263$). \textbf{Left:} Seen scene, \textbf{Right:} New scene. }
    \label{fig:performance_on_kitti}
\end{figure}

\begin{figure}[t]
    \centering
    \includegraphics[width=0.49\linewidth]{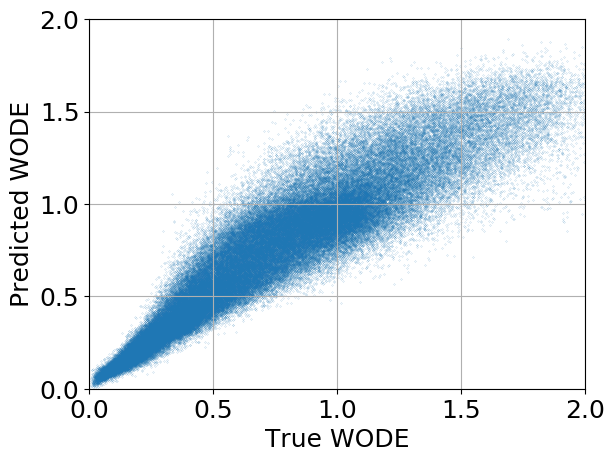}
    \includegraphics[width=0.49\linewidth]{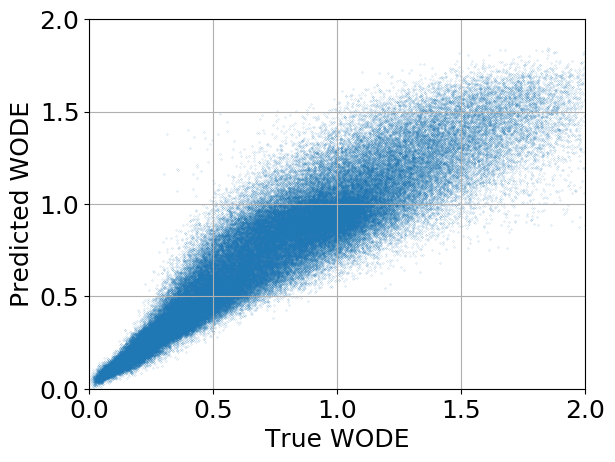}
    \caption{Network performance on EuRoC dataset ($\delta_{thr} = 0.0243$). \textbf{Left:} Seen scene, \textbf{Right:} New scene. }
    \label{fig:peformance_on_euroc}
\end{figure}

The performance of CalQNet is first evaluated on image pairs which were used for training, but with new random disturbances (seen scene).
Secondly, new images from the test set are evaluated with random disturbances (new scene).
As shown in Figures~\ref{fig:performance_on_kitti} and~\ref{fig:peformance_on_euroc} CalQNet is able to capture the general trend for the sensor setup, even from scenes that the network has never encountered before, though performing not as well as on the seen scenes.

In addition, CalQNet appears to perform worse in both cases when \ac{WODE} is extremely high.
This performance drop could be caused by the training disturbances which are generated randomly from a uniform distribution between $\pm1.5d_{thr}$, result in a non-uniformly distributed $\bar{\delta}$ with larger values being less frequent.
The second possible reason is that the cropping and resizing introduced in Section~\ref{ssec:dataset} result in a worse image quality with larger disturbances.
Moreover, motion distortion and blur, although present in the data captured in a moving vehicle, are not explicitly addressed by the analysis, and therefore the results also account for them.
Although the network structure of CalQNet is relatively simple, and potential performance could be gained by using more modern architectures, the experiments demonstrate the sensitivity of CalQNet to artifacts caused by wrong calibrations.

%However, more variability is observed on EuRoC comparing to KITTI. One possible factor contributing to this is the grayscale lower resolution images in EuRoC compared to the more informative RGB images from KITTI.
%The stereo rectification method~\cite{opencv_library} used, however, is only well defined for those two cases leading to inaccuracies and unexpected behavior for extreme cases.

\subsection{WODE and Calibration Parameters}
\label{ssec:eval-wode}
\begin{figure}[!t]
    \centering
    \begin{subfigure}[b]{0.49\linewidth}
        \centering
        \includegraphics[height=0.65\linewidth]{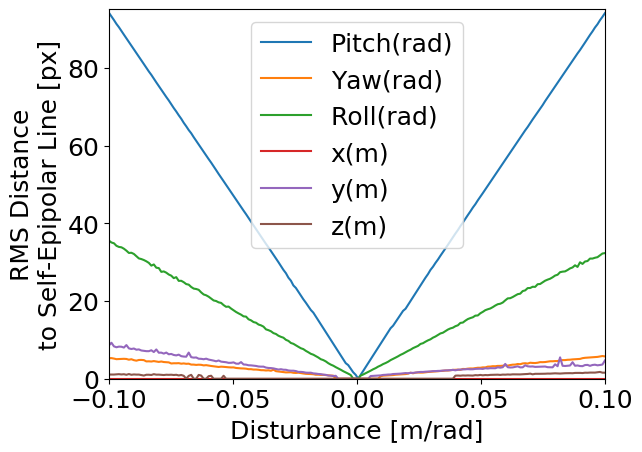}
    \end{subfigure}
    \begin{subfigure}[b]{0.49\linewidth}
        \centering
        \includegraphics[height=0.65\linewidth]{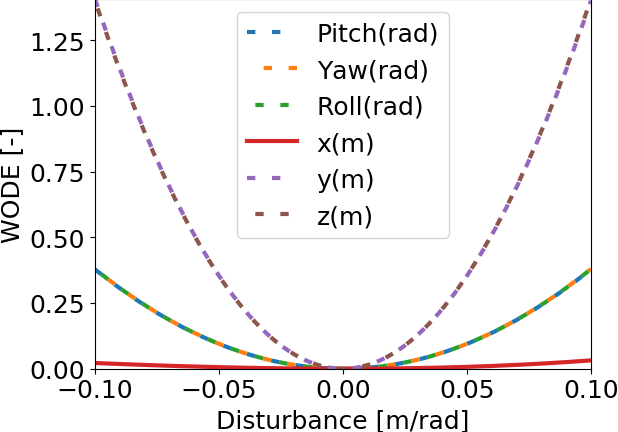}
    \end{subfigure}

    \caption{Evaluation of the influence of disturbances in different \ac{dof} based on an image from KITTI \textbf{Left}: $e_{epi}$; \textbf{Right} \ac{WODE}.} 
    \label{fig:wode_eval_kitti}
\end{figure}

\begin{figure}[!t]
    \centering
    \begin{subfigure}[b]{0.49\linewidth}
        \centering
        \includegraphics[height=0.65\linewidth]{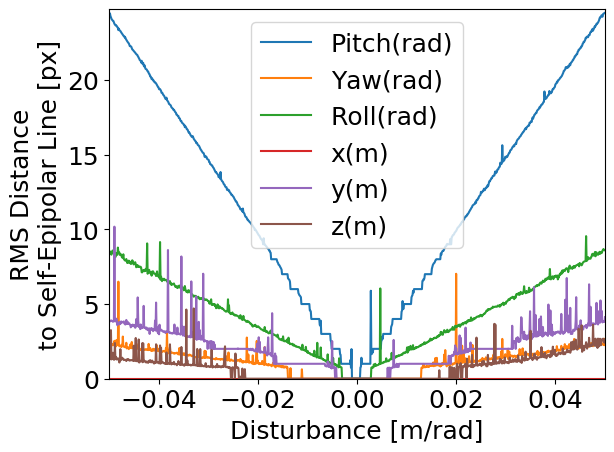}
    \end{subfigure}
    \begin{subfigure}[b]{0.49\linewidth}
        \centering
        \includegraphics[height=0.65\linewidth]{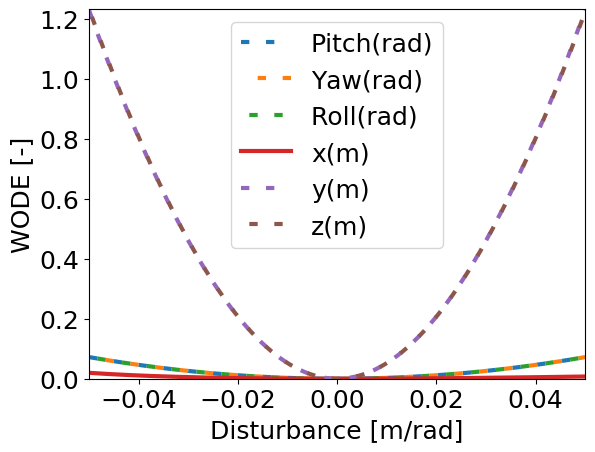}
    \end{subfigure}

    \caption{Evaluation of the influence of disturbances in different \ac{dof} based on a an image from EuRoC \textbf{Left}: $e_{epi}$, discontinuities result from SIFT detection/matching failures; \textbf{Right} \ac{WODE}.} 
    \label{fig:wode_eval_euroc}
\end{figure}

The most common way to evaluate stereo rectification related algorithms is by utilizing epipolar geometry.
Thus, it is of interest to compare our metric with such an approach.
We calculate the root mean squared distances $e_{epi}$ of matched SIFT features~\cite{Lowe1999ObjectFeatures} in a stereo image to their corresponding epipolar lines.
The influences on \ac{WODE} and $e_{epi}$ from disturbances in different degrees of freedom are shown in Figures~\ref{fig:wode_eval_kitti} and~\ref{fig:wode_eval_euroc}.
Calculations are based on the KITTI \texttt{2011-09-26 calibration session} and EuRoC \texttt{MH\textunderscore{01}\textunderscore{easy}} sequences.

\begin{figure}[!t]
    \centering
    \begin{subfigure}[b]{0.49\linewidth}
        \centering
        \includegraphics[height=0.65\linewidth]{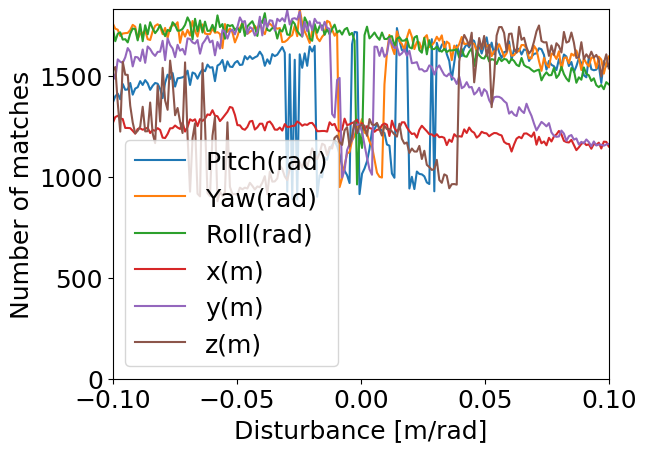}
    \end{subfigure}
    \begin{subfigure}[b]{0.49\linewidth}
        \centering
        \includegraphics[height=0.65\linewidth]{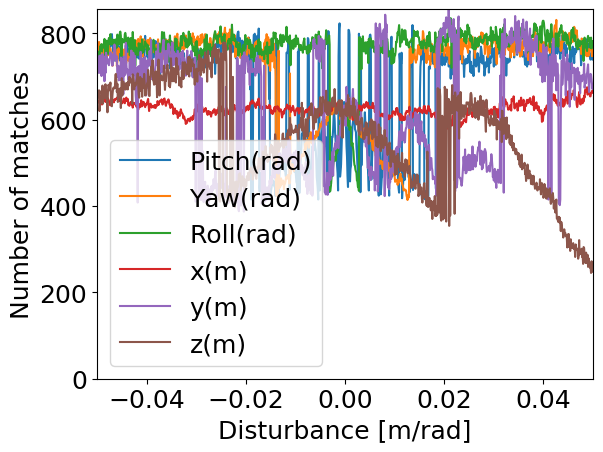}
    \end{subfigure}

    \caption{Number of successfully matched SIFT features w.r.t. disturbances on different \ac{dof} on \textbf{Left}: an image from KITTI; \textbf{Right:} an image from EuRoC.} 
    \label{fig:sift_number}
\end{figure}

Both metrics grow as the extrinsic parameters diverge from the provided calibration.
This means that for both metrics larger scores represent larger disturbances.
However, it can be observed in Figures~\ref{fig:wode_eval_kitti} and~\ref{fig:wode_eval_euroc} that $e_{epi}$ has more discontinuities and is less smooth than the analytically calculated \ac{WODE}.
This is mainly due to the fact that the number of features successfully detected and matched decreases when disturbances increase and even fails to detect any features after a certain point, as shown in Figure~\ref{fig:sift_number}.

Furthermore, disturbances in the same category (translation, rotation) have similar influence on \ac{WODE}, with the exception of the direction of the baseline where the influence on \ac{WODE} is minimal.
This further helps \ac{WODE} to generalize in capturing the calibration quality when all \acp{dof} are disturbed, as they have a more balanced effect on the total miscalibration metric.

%%%%%%%%%%%%%%%%%%%%%%%%%%%%%%%%%%%%%%%%%%%%%%%%%%%%%%%%%%%%%%%%
%%%%%%%%%%%%%%%%%%%%% Section 4.3 %%%%%%%%%%%%%%%%%%%%%%%%%%%%%%
%%%%%%%%%%%%%%%%%%%%%%%%%%%%%%%%%%%%%%%%%%%%%%%%%%%%%%%%%%%%%%%%
\subsection{WODE and Odometry Algorithm Performance}
\label{sec:eval-orbslam}
\begin{figure}[!t]
    \centering
    \begin{subfigure}[b]{0.49\linewidth}
        \centering
        \includegraphics[width=\linewidth]{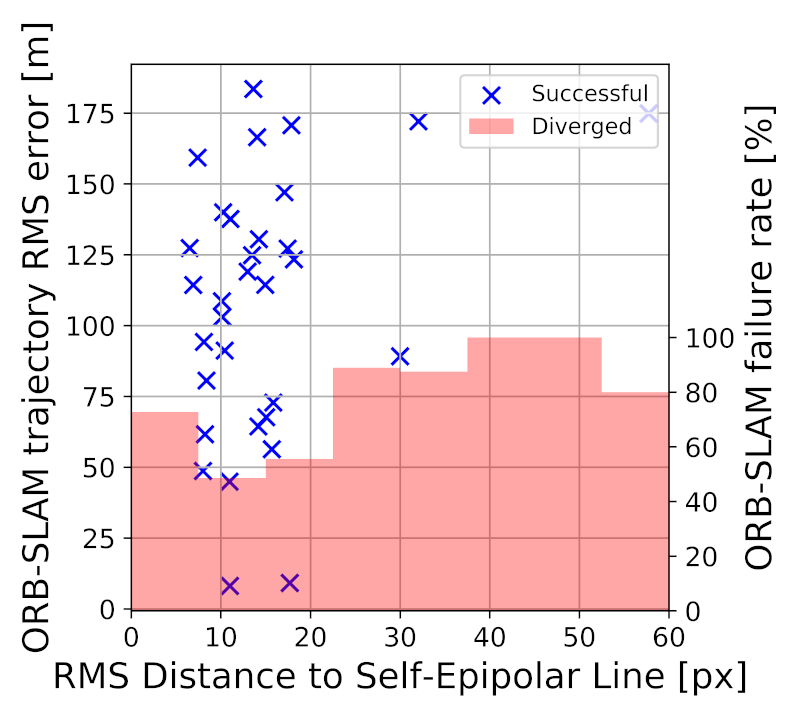}
    \end{subfigure}
    \begin{subfigure}[b]{0.49\linewidth}
        \centering
        \includegraphics[width=\linewidth]{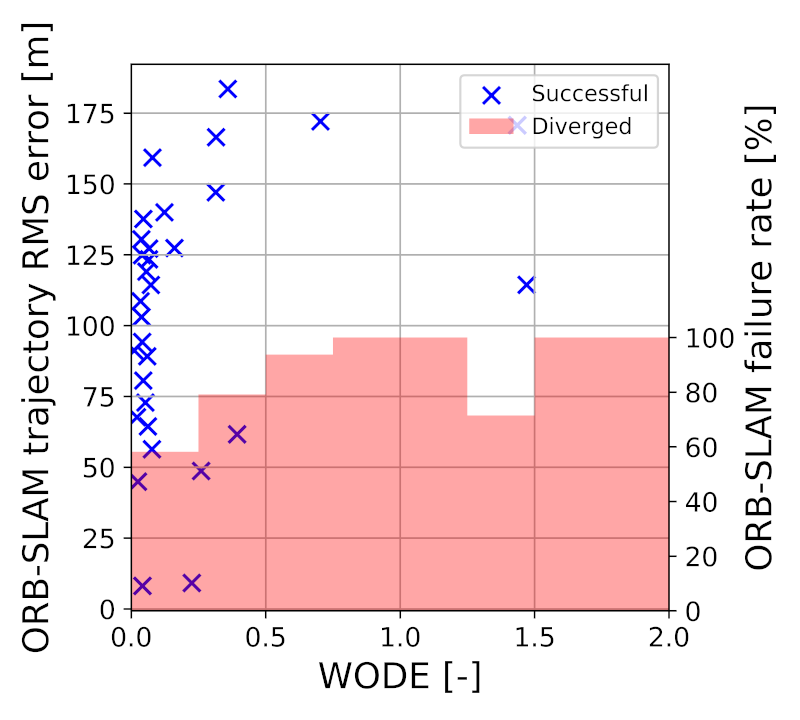}
    \end{subfigure}
    \caption{Performance of ORB-SLAM2 evaluated on KITTI \texttt{2011\textunderscore{09}\textunderscore{30}\textunderscore{drive}\textunderscore{0034}} (total trajectory length $\unit[920]{m}$) w.r.t \textbf{Left:} $e_{epi}$; \textbf{Right:} \ac{WODE}.}
    \label{fig:drift_kitti}
\end{figure}

\begin{figure}[!t]
    \centering
    \begin{subfigure}[b]{0.49\linewidth}
        \centering
        \includegraphics[width=\linewidth]{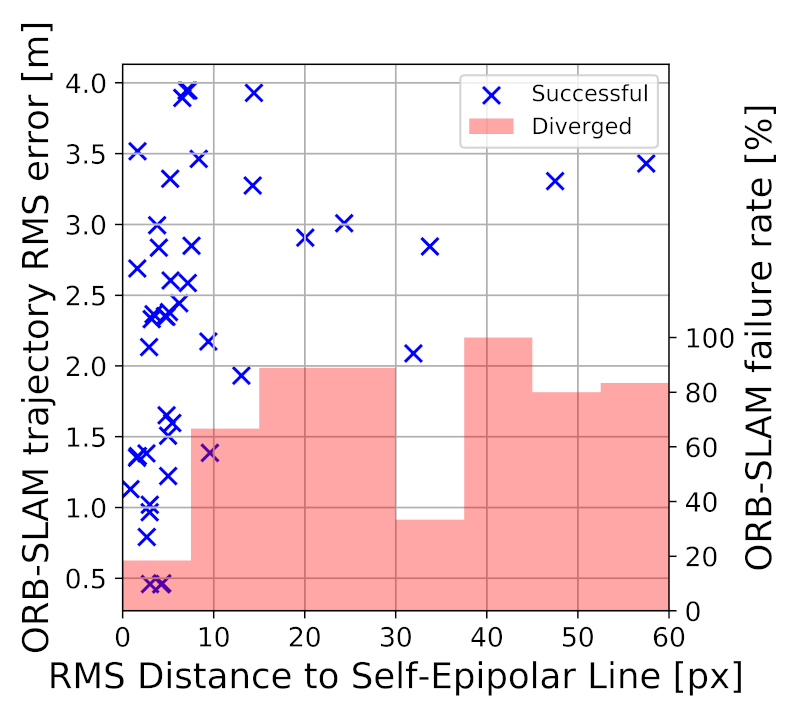}
    \end{subfigure}
    \begin{subfigure}[b]{0.49\linewidth}
        \centering
        \includegraphics[width=\linewidth]{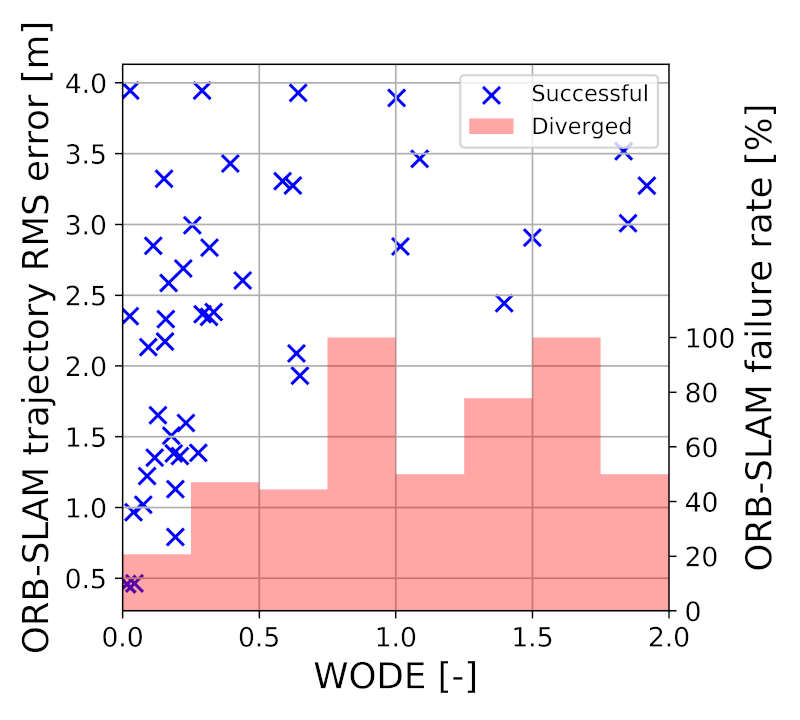}
    \end{subfigure}
    \caption{Performance of ORB-SLAM2 evaluated on EuRoC \texttt{MH\textunderscore{01}\textunderscore{easy}} (total trajectory length $\unit[80]{m}$) w.r.t \textbf{Left:} $e_{epi}$; \textbf{Right:} \ac{WODE}.}
    \label{fig:drift_euroc}
\end{figure}

One way to quantify the calibration quality, from a deployment perspective in a real system, is by evaluating how much the error in calibration influences the task performance, e.g. errors in SLAM or odometry.
Hence, we conducted experiments with ORB-SLAM2~\cite{Mur-Artal2017ORB-SLAM2:Cameras} on both the KITTI and EuRoC datasets, in order to better evaluate the newly proposed metric \ac{WODE}.
Images are rectified with randomly disturbed extrinsic parameters in all 6 \ac{dof}.
Trajectories extracted by the odometry algorithm are compared with the provided ground truth using the trajectory evaluation toolbox~\cite{Zhang2018AOdometry}.
Figure~\ref{fig:drift_kitti} and Figure~\ref{fig:drift_euroc} show the relationship between both $e_{epi}$ and \ac{WODE} w.r.t. the \ac{rmse} of the trajectory estimate from ORB-SLAM2 on KITTI and EuRoC, respectively, as well as the success rate of ORB-SLAM2.
\textit{Successful} is defined as ORB-SLAM2 finishing odometry on a given trajectory regardless of the drift.
Thus, in a \textit{non-successful} or \textit{diverged} run, ORB-SLAM2 loses track and can not continue its operation.

Based on Figures~\ref{fig:performance_on_kitti} and~\ref{fig:peformance_on_euroc}, which show the prediction accuracy of CalQNet for KITTI and EuRoC, we can calculate  a standard deviation of $\sigma_{\text{KITTI}}=0.130$ and $\sigma_{\text{EuRoC}}=0.113$ for unseen scenes on KITTI and Euroc, respectively.
Assuming a two $\sigma$ confidence interval for predicting WODE, and an arbitrarily chosen operating point, for example an \ac{rmse} drift of no more than $\unit[75]{m}$ for KITTI and $\unit[2]{m}$ for EuRoC, respectively, we can now calculate from Figure~\ref{fig:drift_kitti} and Figure~\ref{fig:drift_euroc} how well we can predict miscalibration in these conditions.
In this case the true positive rate will be the number of successful runs of ORB-SLAM2, where the drift is lower than the threshold and the WODE value is withing the confidence interval around zero (meaning we predict the calibration remains unchanged).
Similarly true negatives are all the runs that diverge or have a drift higher than the threshold and also have a WODE value that is further from zero than the confidence interval of prediction (meaning we predict the calibration has changed).
Therefore, for KITTI we have an accuracy of $\unit[62]{\%}$ on predicting how likely we are to maintain our target performance given the current calibration.
As an additional advantage, our method has a very high precision of $\unit[99]{\%}$, meaning that we are very unlikely to trigger a recalibration unnecessarily, i.e. reporting a false positive.
Similarly for EuRoC, for an arbitrary operating point of $\unit[2]{m}$ \ac{rmse} and the same confidence interval of two $\sigma$, we achieve an accuracy of $86\%$ and a precision of $97\%$.

Not taking the failure cases into account, in Table~\ref{tab:corr_kitti} we calculate the Spearman's rank correlation coefficient between ORB-SLAM2 drift and the $e_{epi}$ and WODE metrics presented in Figure~\ref{fig:drift_kitti} and Figure~\ref{fig:drift_euroc}.
On KITTI, only \ac{WODE} shows a significant correlation with the \ac{rmse}.
On EuRoC, both $e_{epi}$ and \ac{WODE} have a significant moderate correlation with the \ac{rmse}, but the correlation with \ac{WODE} is stronger.
This shows that \ac{WODE} is a better metric than $e_{epi}$ to measure the expected performance of a stereo visual odometry algorithm.
Overall, small WODE values correlate not only to a small trajectory error with ORB-SLAM2, but also to a higher success rate regarding the odometry diverging.
The prediction accuracy and confidence could further be increased by using a sequence of predictions from different images to obtain a final decision.

\begin{table}[t]
\vspace{1.5em}
\centering
\begin{tabular}{c c c c c}
\toprule
& \multicolumn{2}{c}{$p$} & \multicolumn{2}{c}{$r$} \\ \cmidrule(lr){2-3} \cmidrule(lr){4-5}
 & EuRoC & KITTI & EuRoC & KITTI  \\
% \cmidrule{2-3} & \cmidrule{4-5}
%  & $p_{EuRoC}$ & $p_{KITTI}$ & $r_{EuRoC}$ & $r_{KITTI}$  \\
\cmidrule{1-5}
$e_{epi}$ & 0.00013 & 0.24 & 0.54 & 0.21 \\
\addlinespace
WODE      & 0.00002 & 0.01 & 0.59 & 0.44 \\ 
\bottomrule
\end{tabular}
\caption{P-value~($p$) and Spearman's rank correlation coefficient~($r$) between ORB-SLAM2 drift (ignoring failure cases) and the different metrics on the KITTI and EuRoC datasets}
\label{tab:corr_kitti}
\end{table}

\section{Conclusion}
\label{sec:conclusion}
Our newly proposed metric \acf{WODE} successfully captures the degree of distortion from extrinsic miscalibration without depending on explicit traditional feature extraction and matching.
We have exploited the semi-synthetic data generation introduced by~\cite{Cramariuc2020LearningDetection} and successfully trained a simple neural network architecture to regress \ac{WODE} from a single stereo image.
In two real-world experiments covering both autonomous driving and micro aerial vehicle applications, we show that CalQNet is able to predict the divergence of ORB-SLAM2~\cite{Mur-Artal2017ORB-SLAM2:Cameras} due to a bad calibration with high confidence and an accuracy between $62\%$ and $86\%$.
CalQNet can be deployed on-board a mobile robotic platform, where it could run at a very low frequency online to prevent potential degradation of a stereo camera system's calibration, due to mechanical wear or other external factors.
Currently, the network only performs the detection based on a single frame, which may cause inconsistency among consecutive frames. 
Further improvement is possible by introducing sequences and using the recurrent information.

\begin{acronym}
\acro{WODE}{\textit{weighted overall disturbance effect}}
\acro{CNN}{convolutional neural network}
\acro{dof}[DoF]{degree of freedom}
\acrodefplural{dof}{degrees of freedom}
\acro{fov}[FoV]{field of view}
\acro{rmse}[RMSE]{root mean squared error}
\end{acronym}

\bibliographystyle{IEEEtran}
\bibliography{references}

\begin{thebibliography}{10}
\providecommand{\url}[1]{#1}
\csname url@rmstyle\endcsname
\providecommand{\newblock}{\relax}
\providecommand{\bibinfo}[2]{#2}
\providecommand\BIBentrySTDinterwordspacing{\spaceskip=0pt\relax}
\providecommand\BIBentryALTinterwordstretchfactor{4}
\providecommand\BIBentryALTinterwordspacing{\spaceskip=\fontdimen2\font plus
\BIBentryALTinterwordstretchfactor\fontdimen3\font minus
  \fontdimen4\font\relax}
\providecommand\BIBforeignlanguage[2]{{%
\expandafter\ifx\csname l@#1\endcsname\relax
\typeout{** WARNING: IEEEtran.bst: No hyphenation pattern has been}%
\typeout{** loaded for the language `#1'. Using the pattern for}%
\typeout{** the default language instead.}%
\else
\language=\csname l@#1\endcsname
\fi
#2}}

\bibitem{strecha2008benchmarking}
C.~Strecha, W.~Von~Hansen, L.~Van~Gool, P.~Fua, and U.~Thoennessen, ``On
  benchmarking camera calibration and multi-view stereo for high resolution
  imagery,'' in \emph{2008 IEEE Conference on Computer Vision and Pattern
  Recognition}.\hskip 1em plus 0.5em minus 0.4em\relax Ieee, 2008, pp. 1--8.

\bibitem{Roth1987AnCalibration}
Z.~S. Roth, B.~W. Mooring, and B.~Ravani, ``{An Overview of Robot
  Calibration},'' \emph{IEEE Journal on Robotics and Automation}, vol.~3,
  no.~5, pp. 377--385, 1987.

\bibitem{Zhang2000ACalibration}
Z.~Zhang, ``{A flexible new technique for camera calibration},'' \emph{IEEE
  Transactions on Pattern Analysis and Machine Intelligence}, vol.~22, no.~11,
  pp. 1330--1334, 2000.

\bibitem{abraham2005fish}
S.~Abraham and W.~F{\"o}rstner, ``Fish-eye-stereo calibration and epipolar
  rectification,'' \emph{ISPRS Journal of photogrammetry and remote sensing},
  vol.~59, no.~5, pp. 278--288, 2005.

\bibitem{rehder2016extending}
J.~Rehder, J.~Nikolic, T.~Schneider, T.~Hinzmann, and R.~Siegwart, ``Extending
  kalibr: Calibrating the extrinsics of multiple imus and of individual axes,''
  in \emph{2016 IEEE International Conference on Robotics and Automation
  (ICRA)}.\hskip 1em plus 0.5em minus 0.4em\relax IEEE, 2016, pp. 4304--4311.

\bibitem{furukawa2009accurate}
Y.~Furukawa and J.~Ponce, ``Accurate camera calibration from multi-view stereo
  and bundle adjustment,'' \emph{International Journal of Computer Vision},
  vol.~84, no.~3, pp. 257--268, 2009.

\bibitem{dang2009continuous}
T.~Dang, C.~Hoffmann, and C.~Stiller, ``Continuous stereo self-calibration by
  camera parameter tracking,'' \emph{IEEE Transactions on Image Processing},
  vol.~18, no.~7, pp. 1536--1550, 2009.

\bibitem{zhang1996motion}
Z.~Zhang, Q.-T. Luong, and O.~Faugeras, ``Motion of an uncalibrated stereo rig:
  self-calibration and metric reconstruction,'' \emph{IEEE Transactions on
  Robotics and Automation}, vol.~12, no.~1, pp. 103--113, 1996.

\bibitem{horaud2000stereo}
R.~Horaud, G.~Csurka, and D.~Demirdijian, ``Stereo calibration from rigid
  motions,'' \emph{IEEE Transactions on Pattern Analysis and Machine
  Intelligence}, vol.~22, no.~12, pp. 1446--1452, 2000.

\bibitem{Visinsky1994RoboticSurvey}
M.~L. Visinsky, J.~R. Cavallaro, and I.~D. Walker, ``{Robotic fault detection
  and fault tolerance: A survey},'' \emph{Reliability Engineering and System
  Safety}, vol.~46, no.~2, pp. 139--158, 1994.

\bibitem{Ni2009SensorTypes}
K.~Ni, N.~Ramanathan, M.~N.~H. Chehade, L.~Balzano, S.~Nair, S.~Zahedi,
  E.~Kohler, G.~Pottie, M.~Hansen, and M.~Srivastava, ``{Sensor network data
  fault types},'' \emph{ACM Transactions on Sensor Networks}, vol.~5, no.~3,
  pp. 1--29, 2009.

\bibitem{Cramariuc2020LearningDetection}
A.~Cramariuc, A.~Petrov, R.~Suri, M.~Mittal, R.~Siegwart, and C.~Cadena,
  ``{Learning Camera Miscalibration Detection},'' \emph{ICRA}, 2020.

\bibitem{faugeras1993three}
O.~Faugeras and O.~A. FAUGERAS, \emph{Three-dimensional computer vision: a
  geometric viewpoint}.\hskip 1em plus 0.5em minus 0.4em\relax MIT press, 1993.

\bibitem{Mur-Artal2017ORB-SLAM2:Cameras}
R.~Mur-Artal and J.~D. Tardos, ``{ORB-SLAM2: An Open-Source SLAM System for
  Monocular, Stereo, and RGB-D Cameras},'' \emph{IEEE Transactions on
  Robotics}, vol.~33, no.~5, pp. 1255--1262, 2017.

\bibitem{Hartley2004MultipleVision}
R.~Hartley and A.~Zisserman, \emph{{Multiple View Geometry in Computer
  Vision}}, 2nd~ed.\hskip 1em plus 0.5em minus 0.4em\relax Cambridge: Cambridge
  University Press, 2004.

\bibitem{Atcheson2010}
B.~Atcheson, F.~Heide, and W.~Heidrich, ``{CALTag: High precision fiducial
  markers for camera calibration},'' \emph{VMV 2010 - Vision, Modeling and
  Visualization}, pp. 41--48, 2010.

\bibitem{Zhang2018AOdometry}
Z.~Zhang and D.~Scaramuzza, ``{A Tutorial on Quantitative Trajectory Evaluation
  for Visual(-Inertial) Odometry},'' \emph{IEEE International Conference on
  Intelligent Robots and Systems}, pp. 7244--7251, 2018.

\bibitem{Schneider2019Observability-AwareEstimation}
T.~Schneider, M.~Li, C.~Cadena, J.~Nieto, and R.~Siegwart,
  ``{Observability-Aware Self-Calibration of Visual and Inertial Sensors for
  Ego-Motion Estimation},'' \emph{IEEE Sensors Journal}, vol.~19, no.~10, pp.
  3846--3860, 2019.

\bibitem{Mendoza2012}
\BIBentryALTinterwordspacing
J.~P. Mendoza and R.~Simmons, ``{Mobile Robot Fault Detection based on
  Redundant Information Statistics},'' \emph{2012 International Conference on
  Intelligent Robots and Systems}, 2012. [Online]. Available:
  \url{http://www.cs.cmu.edu/~mmv/papers/12irosw-MendozaVelosoSimmons.pdf}
\BIBentrySTDinterwordspacing

\bibitem{Roumeliotis1998SensorRobot}
S.~I. Roumeliotis, G.~S. Sukhatme, and G.~A. Bekey, ``{Sensor fault detection
  and identification in a mobile robot},'' in \emph{IEEE International
  Conference on Intelligent Robots and Systems}, vol.~3, no. October, 1998, pp.
  1383--1387.

\bibitem{Sundvall2006}
P.~Sundvall and P.~Jensfelt, ``{Fault detection for mobile robots using
  redundant positioning systems},'' \emph{Proceedings - IEEE International
  Conference on Robotics and Automation}, vol. 2006, no. May, pp. 3781--3786,
  2006.

\bibitem{bar-shalom}
Y.~Bar-Shalom, T.~Kirubarajan, and X.-R. Li, \emph{Estimation with Applications
  to Tracking and Navigation}.\hskip 1em plus 0.5em minus 0.4em\relax USA: John
  Wiley \& Sons, Inc., 2002.

\bibitem{Geiger2013}
A.~Geiger, P.~Lenz, C.~Stiller, and R.~Urtasun, ``{Vision meets robotics: The
  KITTI dataset. The International Journal of Robotics Research},'' \emph{The
  International Journal of Robotics Research}, no. October, pp. 1--6, 2013.

\bibitem{BROMLEY1993SIGNATURENETWORK}
J.~Bromley, J.~W. Bentz, L.~Bottou, I.~Guyon, Y.~Lecun, C.~Moore,
  E.~S{\"{A}}CKINGER, and R.~Shah, ``{Signature verification using a "SIAMESE"
  time delay neural network},'' \emph{International Journal of Pattern
  Recognition and Artificial Intelligence}, vol.~07, no.~04, pp. 669--688, 8
  1993.

\bibitem{Kingma2015Adam:Optimization}
D.~P. Kingma and J.~L. Ba, ``{Adam: A method for stochastic optimization},'' in
  \emph{3rd International Conference on Learning Representations, ICLR 2015 -
  Conference Track Proceedings}, 2015, pp. 1--15.

\bibitem{Srivastava2014Dropout:Overfitting}
N.~Srivastava, G.~Hinton, A.~Krizhevsky, I.~Sutskever, and R.~Salakhutdinov,
  ``{Dropout: A simple way to prevent neural networks from overfitting},''
  \emph{Journal of Machine Learning Research}, vol.~15, pp. 1929--1958, 2014.

\bibitem{Glorot2010UnderstandingNetworks}
X.~Glorot and Y.~Bengio, ``{Understanding the difficulty of training deep
  feedforward neural networks},'' in \emph{Journal of Machine Learning
  Research}, vol.~9, 2010, pp. 249--256.

\bibitem{Burri2016TheDatasets}
M.~Burri, J.~Nikolic, P.~Gohl, T.~Schneider, J.~Rehder, S.~Omari, M.~W.
  Achtelik, and R.~Siegwart, ``{The EuRoC micro aerial vehicle datasets},''
  \emph{International Journal of Robotics Research}, vol.~35, no.~10, pp.
  1157--1163, 2016.

\bibitem{Lowe1999ObjectFeatures}
D.~G. Lowe, ``{Object Recognition from Local Scale-Invariant Features},'' Tech.
  Rep., 1999.

\end{thebibliography}

\end{document}